\documentclass[sigconf]{acmart}
\AtBeginDocument{%
  }

\usepackage{float} 
\usepackage{multirow}
\usepackage{array}
\usepackage{makecell} 
\usepackage{graphicx} 
\usepackage{algorithm}
\usepackage{algpseudocode}
\usepackage{amsmath}
\usepackage{mathtools}
\usepackage{placeins}

\usepackage{etoolbox}
\makeatletter
\renewcommand{\@fnsymbol}[1]{\ensuremath{\ast}} 
\makeatother
\usepackage{boldline, hhline}
\usepackage{colortbl}  
\definecolor{mygray}{RGB}{230,230,230}
\definecolor{blue}{RGB}{229,233,243}
\definecolor{bluegray}{RGB}{241,244,250}
\definecolor{orangegray}{RGB}{251,244,242}
\definecolor{orange}{RGB}{248,231,228}
\definecolor{gree}{RGB}{226,239,213}
\definecolor{greegray}{RGB}{245,249,241}
\definecolor{tab1huang}{RGB}{255,240,193}

\setcopyright{acmlicensed}
\copyrightyear{2025}
\acmYear{2025}
\acmConference[MM '25]{In Proceedings of the 33rd ACM International Conference on Multimedia}{October 27--31, 2025}{Dublin, Ireland}

\begin{document}

\title{Positive Style Accumulation: A Style Screening and Continuous Utilization Framework for Federated DG-ReID}

\author{Xin Xu}
\affiliation{%
  \institution{Wuhan University of \\ Science and Technology}
  \city{Wuhan}
  \country{China}}
\email{xuxin@wust.edu.cn}

\author{Chaoyue Ren}
\affiliation{%
  \institution{Wuhan University of \\ Science and Technology}
  \city{Wuhan}
  \country{China}}
\email{renchaoyue@wust.edu.cn}

\author{Wei Liu}
\authornote{Corresponding author}
\affiliation{%
  \institution{Wuhan University of \\ Science and Technology}
  \city{Wuhan}
  \country{China}}
\email{liuwei@wust.edu.cn}

\author{Wenke Huang}
\authornotemark[1]
\affiliation{%
 \institution{Wuhan University}
 \city{Wuhan}
 \country{China}}
\email{wenkehuang@whu.edu.cn}

\author{Bin Yang}
\affiliation{%
  \institution{Wuhan University}
  \city{Wuhan}
  \country{China}}
\email{yangbin_cv@whu.edu.cn}

\author{Zhixi Yu}
\affiliation{%
  \institution{Wuhan University of \\ Science and Technology}
  \city{Wuhan}
  \country{China}}
\email{yuzhixi@wust.edu.cn}

\author{Kui Jiang}
\affiliation{%
  \institution{Harbin Institute of Technology}
  \city{Harbin}
  \country{China}}
\email{jiangkui@hit.edu.cn}

\renewcommand{\shortauthors}{Xin Xu, Chaoyue Ren, Wei Liu, Wenke Huang, Bin Yang, Zhixi Yu, Kui Jiang}

\begin{abstract}
  The Federated Domain Generalization for Person re-identification (FedDG-ReID) aims to learn a global server model that can be effectively generalized to source and target domains through distributed source domain data. Existing methods mainly improve the diversity of samples through style transformation, which to some extent enhances the generalization performance of the model. However, we discover that \textbf{not all styles contribute to the generalization performance}. Therefore, we define styles that are \textbf{beneficial/harmful} to the model’s generalization performance as \textbf{positive/negative styles}. Based on this, new issues arise: \textit{\textbf{How to effectively screen and continuously utilize the positive styles.}} To solve these problems, we propose a \textbf{S}tyle \textbf{S}creening and \textbf{C}ontinuous \textbf{U}tilization \textbf{(SSCU)} framework. Firstly, we design a Generalization Gain-guided Dynamic Style Memory (GGDSM) for each client model to screen and accumulate generated positive styles. Specifically, the memory maintains a prototype initialized from raw data for each category, then screens positive styles that enhance the global model during training, and updates these positive styles into the memory using a momentum-based approach. Meanwhile, we propose a style memory recognition loss to fully leverage the positive styles memorized by GGDSM. Furthermore, we propose a Collaborative Style Training (CST) strategy to make full use of positive styles. Unlike traditional learning strategies, our approach leverages both newly generated styles and the accumulated positive styles stored in memory to train client models on two distinct branches. This training strategy is designed to effectively promote the rapid acquisition of new styles by the client models, ensuring that they can quickly adapt to and integrate novel stylistic variations. Simultaneously, this strategy guarantees the continuous and thorough utilization of positive styles, which is highly beneficial for the model's generalization performance. Extensive experimental results demonstrate that our method outperforms existing methods in both the source domain and the target domain.
\end{abstract}




\maketitle

\begin{figure}
    \centering
    \includegraphics[width=0.47\textwidth]{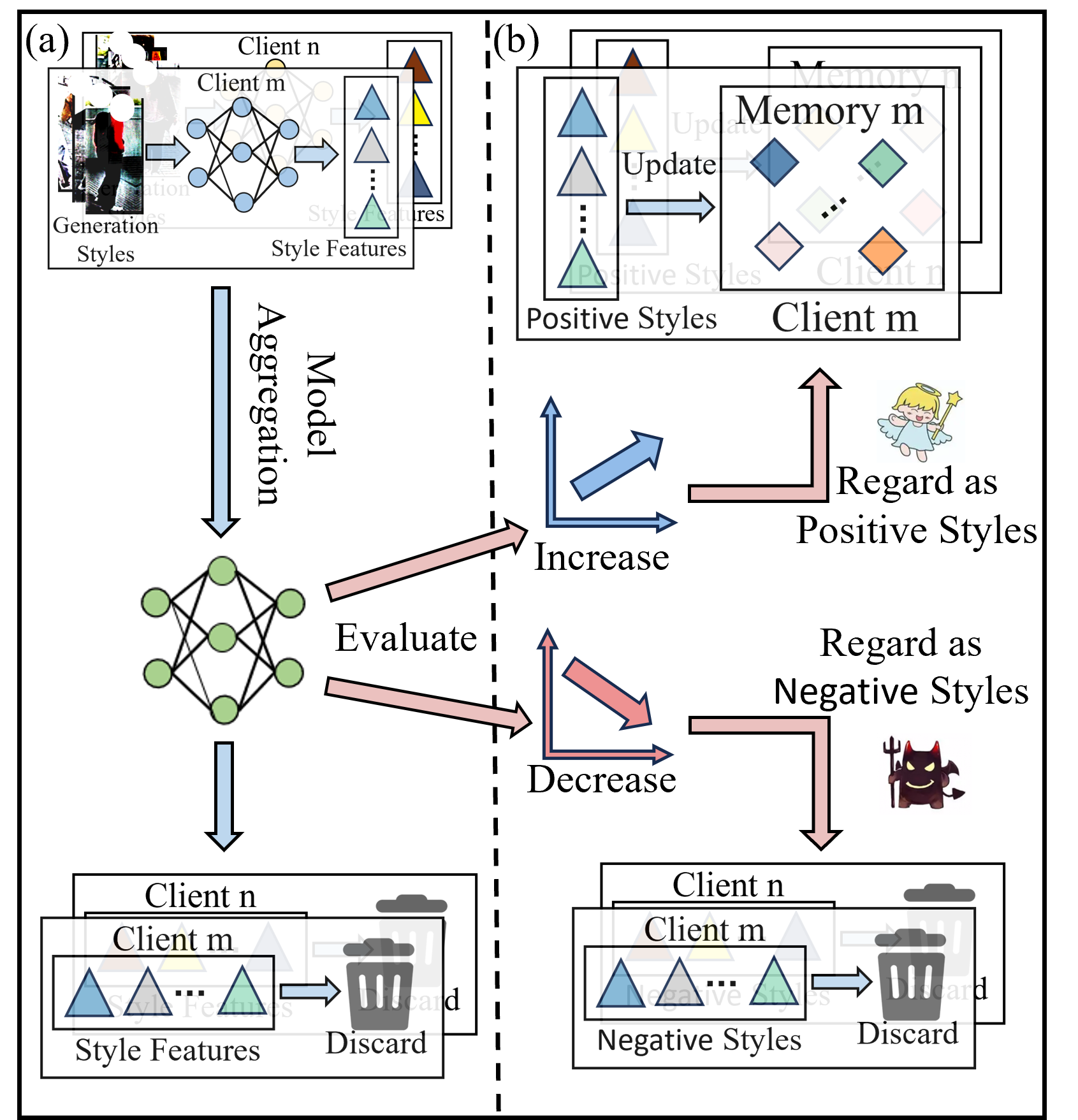}
    \caption{The differences between SSCU and existing methods are as follows: (a) Traditional methods discard the generated styles immediately after each round of training. (b) SSCU evaluates the global model at the end of each round of training, and if the model's performance has improved, the generated styles from the current round are regarded as positive styles and updated into the memory; otherwise, they are considered negative styles and discarded directly.}    
    \label{fig1} 
\end{figure}

\section{Introduction}
In recent years, Person re-identification (ReID) has garnered significant research attention, aiming to achieve accurate cross-camera recognition of the same individual. With the success of deep learning, numerous high-performance ReID methods have been proposed \cite{1,2,3,4,5,6,7,8,9,10}. However, constrained by limitations in data collection and the complexity of real-world scenarios, these methods often underperform when deployed to unseen domains. To address this, recent efforts have focused on Domain Generalization for Person re-identification (DG-ReID) \cite{11,12,13,14,16,17}. DG-ReID aims to train models on multiple source domains and test on unseen target domains, thereby enhancing robust cross-domain generalization. However, existing approaches rely on large-scale centralized labeled datasets, which often raise critical data privacy concerns in practical applications \cite{18,19}.


To solve this issue, federated learning has been introduced into DG-ReID \cite{20,21,22}, termed Federated Domain Generalization for Person re-identification (FedDG-ReID). Federated learning is a distributed machine learning framework \cite{23,GGEUR_CVPR25,FLSurveyandBenchmarkforGenRobFair_TPAMI24} that facilitates knowledge sharing through a cross-device/cross-institution collaborative training paradigm while rigorously safeguarding data privacy. The technology embodies the core principle of "moving models, not data," where participants solely exchange updates to model parameters while retaining raw data locally, thereby effectively addressing the dual challenges of data silos and privacy leakage. However, due to the limited amount of data that each client can access, and the significant heterogeneity between the data of different clients, traditional centralized generalization strategies cannot be directly applied. Existing methods predominantly focus on generating synthetic data via style transfer to simulate unseen domains \cite{25,26,27,28,29,30,FCCLPlus_TPAMI23}. However, as shown in Fig. \ref{fig1},these methods \textit{\textbf{overlook that not all styles contribute to the model's generalization performance}}, thereby lacking the capability to screen and utilize positive styles, which is crucial for the model’s generalization performance in both the source and target domains.


In this paper, we propose a \textbf{S}tyle \textbf{S}creening and \textbf{C}ontinuous \textbf{U}tilization  \textbf{(SSCU)} framework to address the previously outlined issues of \textbf{positive style selection, memory, and continuous utilization}, achieving robust cross-domain generalization while ensuring privacy preservation. Specifically, we design a \underline{G}eneralization \underline{G}ain-guided \underline{D}ynamic \underline{S}tyle \underline{M}emory (GGDSM) for each client to enable selection and cross-round accumulation of positive styles. (1) Initialization: To build a robust identity-discriminative feature representation for each person identity, we perform clustering and averaging of all training data based on identity before the official training starts on each client. category prototypes emphasize consistency within classes and differences between classes, and can be used to guide the model to learn more discriminative feature representations. (2) Positive style selection: At the end of each training round on the client, we evaluate the optimization effect of the generated styles on the global model. Based on the evaluation results, we determine whether these generated styles are positive for the model update, and if they are, we update them to the memory for continuous utilization, otherwise, we consider them to be negative styles and discard them directly. (3) Update strategy: Category prototypes are updated via momentum-based integration, ensuring stable incorporation of new styles while preserving previously memorized positive patterns. This helps the model progressively learn domain-invariant feature extraction capabilities. Furthermore, in order to realize the full use of the style in the memory, we propose a \underline{C}ollaborative \underline{S}tyle \underline{T}raining (CST) training strategy comprising two parallel training branches: (1) New style adaptation branch: In each iteration, new stylized data is randomly generated, and features are extracted using the client-global model for loss calculation. The client-global model downloaded from the server possesses better generalization knowledge, making it more suitable for rapidly learning new style changes within a short period of time. (2) Positive style continuous utilization branch: In this branch, the client-local model and the client-global model are trained using the original images, and then optimized using a loss function based on the dynamic style memory. Since the category prototypes stored in the memory remember all the positive styles from previous rounds, this branch allows the model to continuously make use of them.


Our main contributions can be summarized as follows:

\begin{itemize}
\item \textbf{Empirical Contribution.} We discover that not all styles generated through style transformation methods contribute to the improvement of model generalization performance. Some styles may introduce invalid data, which is instead detrimental to model optimization.

\item \textbf{Framework Contribution.} We propose a style screening and continuous utilization framework that effectively screens, memorizes, and continuously utilizes generative styles beneficial to model generalization performance with minimal additional overhead.

\item \textbf{Technical Contribution.} We propose GGDSM and CST. GGDSM screens and memorizes styles that are positive for model generalization, while CST leverages both newly generated styles and the accumulated positive styles stored in memory to train client models. This enables the models to quickly adapt to new styles and continuously utilize positive styles, resulting in significant improvements in generalization performance in both the source and target domains.

\end{itemize}

\section{Related work}

\subsection{Domain Generalization for Person Re-identification}

Recently, Domain Generalization for person re-identification have gained widespread attention. The goal is to enable models to learn from one or multiple source domains and generalize effectively to unseen domains. The key to domain generalization methods lies in enabling models to learn robust domain-invariant feature representations. Existing approaches achieve this goal from various perspectives, including optimizing training strategies \cite{31,32,33,34,11057929}, leveraging causal mechanisms \cite{35,36}, and employing data augmentation \cite{26,29,27,28,25,30,li2025rethinking}. Recent methods mainly seek to improve DG performance through three aspects: meta-learning, feature disentanglement, and data augmentation.

Methods based on meta-learning aim to enhance model generalization by simulating realistic training-test domain shifts. For example, Choi et al. \cite{16} introduced MetaBIN, which combines Batch Normalization (BN) and Instance Normalization (IN) through a meta-learning training strategy and a set of learnable balancing parameters. Dai et al. \cite{12} proposed a Relation-aware Mixture of Experts (RaMoE) method, integrating meta-learning into a novel mixture-of-experts paradigm via an efficient voting-based fusion mechanism to dynamically aggregate multi-source domain features. 


Some methods adopt feature disentanglement strategies, focusing on disentangling different attribute features of pedestrians to help models learn more discriminative identity features. Jin et al. \cite{41} proposed a Style Normalization and Recovery (SNR) module, which eliminates the influence of style features via IN layers while extracting and restoring identity-discriminative features filtered by IN. Zhang et al. \cite{14} designed an Adaptive Cross-domain Learning (ACL) framework, which introduces multiple parallel feature embedding networks to capture domain-invariant and domain-specific features separately and adaptively aggregates them using domain-aware adapters to mitigate cross-domain interference. Zhang et al. \cite{43} developed a Disentangled Invariant Representation (DIR) framework, constructing a Structural Causal Model (SCM) between identity-specific and domain-specific factors to eliminate spurious correlations and filter domain-related information.


Regarding data augmentation, recent studies have demonstrated that diversifying image styles (e.g., background environments, lighting conditions, camera positions) significantly enhances model generalization. Zhou et al. \cite{29} proposed MixStyle, which generates new styles by linearly combining style feature statistics from different source domains. Nuriel et al. \cite{46} introduced Permuted adaIN (pAdaIN), exchanging feature statistics within a batch to reduce network reliance on global image statistics and enhance the utilization of shape and local image cues.


\subsection{Federated Domain Generalization for Person Re-identification}

Federated learning is a distributed machine learning technique that enables multiple devices to collaborate while ensuring privacy protection \cite{FPL_CVPR23,FCCL_CVPR22} . McMahan et al. \cite{48} proposed the first federated learning algorithm, FedAvg, which aggregates locally trained client models on the server via averaging and redistributes the aggregated model to clients for further training.


In recent years, as privacy concerns have grown in domain generalization tasks, integrating federated learning into ReID frameworks has emerged as a promising research direction. Although FedDG-ReID shares the same objective as DG tasks, traditional DG methods, which rely on large-scale centralized data for training, struggle to adapt directly under privacy-constrained decentralized data architectures. Existing mainstream FedDG-ReID approaches primarily address this challenge through three directions: client model training strategies, server model aggregation strategies, and local training data diversification. For example: Zhuang et al. \cite{49} combined knowledge distillation to transfer client model knowledge to the server with dynamic weight adjustment methods, which dynamically adjust model aggregation weights based on client model variations. Wu et al. \cite{20} introduced a local expert model for each client to enrich local knowledge acquisition and aggregated only feature embedding networks during server updates to preserve local classification knowledge. Zhang et al. \cite{51} optimized local models via proximal and feature regularization terms to improve local training accuracy, while using cosine similarity of backbone features to determine global aggregation weights for each local model, ensuring global convergence. Regarding local training data diversification: Yang et al. \cite{52} designed a Style Transfer Model (STM) for each client to generate new styles for local data, with loss functions constraining style diversity and authenticity. Liu et al. \cite{24} simulated unseen domains via a domain compensation network (DCN), exposing models to broader data distributions during training.


\begin{figure*}[t]
    \centering
    \includegraphics[width=\textwidth]{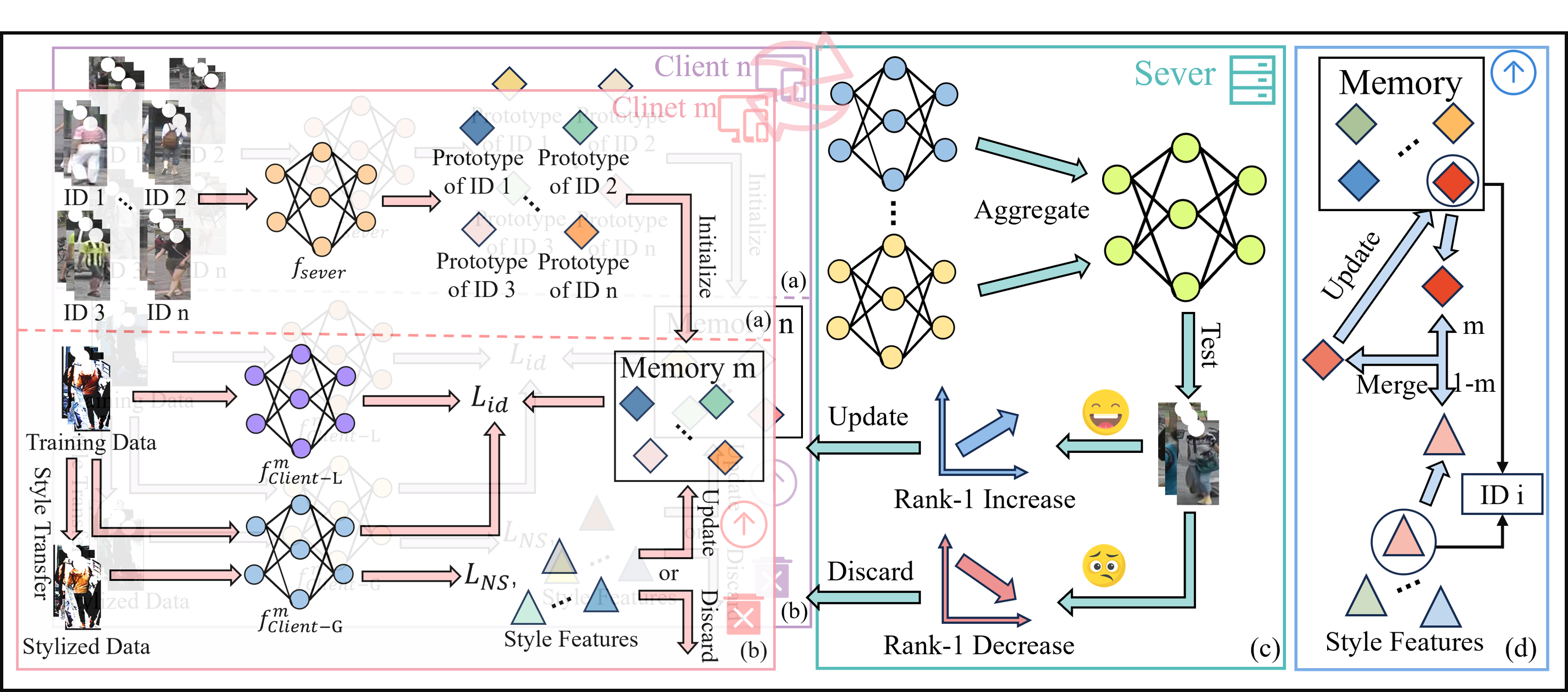}
    \caption{Illustration of our SSCU method. (a) Memory Initialization. Before the training begins, clients use the pretrained global model downloaded from the server to extract features, obtaining the category prototypes for each identity. (b) Our CST strategy. Specifically, the figure illustrates the training process of both the new style adaptation branch and the positive style continuous utilization branch, as well as the generation of style features used for updating the memory. (c) Subsequent learning of our method. After local training concludes, the server aggregates the client-global models uploaded by each client and evaluates them. The evaluation results decides whether to memorize the generated styles of this round. (d) We present the details of memory update. New style features are updated to the category prototypes corresponding to their identities in a momentum-based manner.}    
    \label{fig2}
    \vspace{-2mm}
\end{figure*}
\vspace{-2mm}

\section{Methodology}

For FedDG-ReID, we provide $K$ source domains $D_\text{source} = \left\{ D_k \right\}_{k=1}^{K}$. During training, there is no interaction among the source domain. The goal of the training task is to obtain a global model that can be generalized. During the testing phase, the global model is tested directly in the source or target domain $D_{T}$.


As shown in Fig. \ref{fig2}, our method mainly consists of two parts: (1) generalization gain-guided dynamic style memory (Section~\ref{sec31}) and (2) collaborative style training strategy (Section \ref{sec32}). The collaborative training process under the federated framework mainly consists of five steps: Step (1) Local training: Use specific training strategies to optimize the model located on the client; Step (2) Model upload: Upload the specific model optimized by local training to the server for aggregation; Step (3) Model aggregation: Use specific methods to aggregate the models uploaded by each client to obtain a global model; Step (4) Model redistribution: Redistribute the new global model to each client, and continue to iterate steps (1) to (4) until convergence; Step (5) Test the final global model on the target domain. Our method mainly works in step (1).

\vspace{-3.1mm}

\subsection{Generalization Gain-Guided Dynamic Style Memory}\label{sec31}

Style diversification methods can effectively help models learn more discriminative features. Although existing work has demonstrated their effectiveness in the FedDG-ReID task, these methods overlook the potential issue in practical applications that not all generated styles are beneficial to the gains in model generalization performance. Consequently, the model may fail to form long-term memory of positive styles and filter out negative styles. Considering the aforementioned issues, we propose a generalization gain-guided dynamic style memory. We maintain a separate memory for each client, storing the category prototype of each identity, with no interaction between the memories of different clients. Using category prototypes as classification features, which can simplify feature representation and improve classification efficiency, is highly suitable for complex open-set scenarios like ReID.


\textbf{Memory initialization.} As shown in Fig. \ref{fig2}(a), we use a pretrained global model to extract features from all training samples, and then aggregate these features by identity to calculate the average, obtaining the category prototypes to each identity.


\begin{equation}
    M_k^i = \frac{1}{n_k^i} \sum_{x \in X_k^i} f_{\text{pretrain}}(x),
\label{eq1}
\end{equation}

where $X_k^i$ is the set of all images with identity $i$ in client $k$, with a total number of $n_k^i$, and the category prototype is $M_k^i$.


\textbf{Positive style screening and memory update.} We design an update mechanism for the memory that is guided by the generalization performance gain of the global model, in order to selectively preserve positive generated styles and filter out negative ones. Specifically, after each training round, We assess the global model’s performance, with Rank-1 serving as the key indicator of generalization capability. We then compare these evaluation outcomes with those from the preceding round. If the styles produced in the current round have helped to enhance the model’s generalization performance, we regard them as positive styles and update them to the memory, otherwise, we regard them as negative styles and discard them directly. We update the memory using a momentum-based approach, which operates as follows:


\begin{align}
 \hat{x}_k^i &= f_{\text{T}}^k(x_k^i), \label{eq2}\\
 M_k^i \leftarrow (1 - m) M_k^i &+ m \sum_{x \in B_k^i} \| f_{\text{Client-G}}^k(\hat{x}_k^i) \|_2 \label{eq3},
\end{align}

where $B_{k}^i$  represents all images of identity $i$ in the mini-batch training data at the $k-th$ client, $f_{\text{Client-G}}^k$ is the client-global model of the $k-th$ client,  $f_{\text{T}}^k$ is the style transformation model of the $k-th$ client, $\|\cdot\|_2$ indicates the $L_2$ normalization operation, and $m \in [0,1]$ is a momentum coefficient used to control the update rate of category prototypes. By allowing category prototypes to continuously accumulate positive style features through a momentum-based approach over multiple rounds of training, we can not only achieve sustained cumulative memory of positive styles but also implicitly expand the training samples, so that we can also alleviates the challenges posed by limited training data to a certain extent.


\textbf{Style memory recognition loss.} In the task of FedDG-ReID, cross-entropy loss is often used to help the model learn to distinguish between different individuals. It features low computational complexity, ease of implementation and optimization. Existing methods have also demonstrated its rapid convergence, efficient training, and strong application compatibility. However, due to the characteristic of cross-entropy loss that overly focuses on the correctness of identity classification, the trained models tend to overfit to specific styles and struggle to adapt to data heterogeneity. Considering these drawbacks, we have designed a style memory recognition loss. This is achieved by maximizing the similarity between features and prototypes of the same category, and minimizing the similarity with other categories, thus helping the model learn features that are more robust to noise and style variations. Therefore, by applying cross-entropy loss and recognition loss at different training scenes, the strengths of both methods can complement each other, aiding the model in achieving better generalization capability. The implementation of the loss function we designed is as follows:


Given an image, we pass it into the model $f$ that needs to be optimized to perform forward propagation and obtain features. and then calculate the similarity between it and each prototype in the memory. For a source domain $D_k$ with $P_k$ identities, the specific calculation formula is as follows:


\begin{equation}
    L_{id}(f,x_k^i,M_k^i) = -\log \frac{\exp(\|f(x_k^i)\|^T_2 M_k^i / \tau)}{\sum_{n=1}^{P_k} \exp(\|f(x_k^i)\|^T_2 M_k^n / \tau)}, \label{eq4}
\end{equation}

among them, $x_k^i$ is an image with identity $i$ in domain $D_k$, \( \tau \) is the temperature factor, $exp(\cdot)$ denotes the operation of raising $e$ to a power, and the similarity between the extracted features and the category prototypes is calculated using the dot product.


\definecolor{myblue}{RGB}{196,214,230}
\definecolor{mylightblue}{RGB}{53,160,219} 
\definecolor{myllblue}{RGB}{237,242,241}

\begin{algorithm}[t]
\caption{: SSCU}
\label{alg1}

\begin{algorithmic}[1]
\Require{Client set $K$, sample set of identity $i$ at client $k$: $X_k^i = [ X_k^{(i,1)}, ..., X_k^{(i,n_k^i)} ]$, the total number of pedestrian identities on client $k$: $P_k$, the client-local model of a client $k$: $f_{Client-L}^{k}$ and style transformation model: $f_{T}^{k}$, server model $f_{server}$, communication rounds $E$}
\Function{initialize}{$X_k, M_k, f$}
    \For{$id = 1,2,...,P_k$}
        \State $M_k^{id} \leftarrow \frac{1}{n_k^{id}} \sum_{n=1}^{n_k^{id}} \| f(X_k^{(id,n)}) \|_2 \text{ in Eq.\textcolor{red}{~\ref{eq1}}}
$
    \EndFor
    \State \Return $M_k$
\EndFunction
\Function{update}{$M_k, F_k$}
    \For{$id = 1,2,...,P_k$}
        \State $M_{k}^{id} \leftarrow (1-m) M_k^{id} + m F_k^{id} \text{ in Eq.\textcolor{red}{~\ref{eq3}}}$
        \State \textcolor{mylightblue}{\Comment{$F_k^{id}$ is $L_2$ normalized}}
    \EndFor
    \State \Return $M_k$
\EndFunction
\Function{CST}{$f_{Client-G}^{k}, f_{Client-L}^{k}, f_{T}^{k}, X_k, M_k$}
    \State $B_k \leftarrow \text{Data sampling}(X_k)$ \textcolor{mylightblue}{\Comment{Common random sampling}}
    \State $\hat{B}_k \leftarrow f^{k}_{T}(B_k) \text{ in Eq.\textcolor{red}{~\ref{eq2}}}$ 
    \State $f_{Client-G}^{k} \leftarrow L_{NS}(f_{Client-G}^{k}, \hat{B}_k) \text{ in Eq.\textcolor{red}{~\ref{eq5}}}$
     \State $\begin{aligned}[t]
            & f_{Client-G}^{k}, f_{Client-L}^{k} \leftarrow L_{PS}( \\ 
            & f_{Client-G}^{k}, f_{Client-L}^{k}, B_k, M_k) \text{ in Eq.\textcolor{red}{~\ref{eq8}}} 
            \end{aligned}$
    \State $\hat{F}_k = \|f_{Client-G}^{k}(\hat{B}_k)\|_2$\textcolor{mylightblue}{\Comment{$\hat{F}_k$ is style features}}
    \State \Return $f_{Client-G}^{k}, f_{Client-L}^{k}, \hat{F}_k$
\EndFunction

\For{$k \in K \text{ in parallel}$}
    \State $M_{k} \leftarrow INITIALIZE(X_k, M_k, f_{server}) $
\EndFor
\For{$e = 1,2,...,E$}
    \For{$k \in K \text{ in parallel}$}
        \State $f_{Client-G}^{k} \leftarrow \text{Distribute}(f_{server})$ 
        \State \textcolor{mylightblue}{\Comment{Also known as server model download}}
        \State $\begin{aligned}[t]
            & f_{Client-G}^{k},f_{Client-L}^{k},\hat{F}_k \leftarrow \text{CST}( \\ 
            & f_{Client-G}^k, f_{Client-L}^{k}, f_{T}^{k}, X_k, M_k) 
            \end{aligned}$
    \EndFor
    \State $f_{server} \leftarrow \sum_{k=1}^{K} \frac{N_k}{N} f^k_{\text{Client-G}}$\textcolor{mylightblue}{\Comment{Evaluate after aggregation}}
    \If{$e = 0$ or $f_{server}$'s Rank-1 increase} 
    \State \textcolor{mylightblue}{\Comment{Compare with the last round}}
        \For{$k \in K \text{ in parallel}$}
            \State $M_{k} \leftarrow \text{UPDATE}(M_k,\hat{F}_k) $
        \EndFor
    \EndIf
\EndFor
\State \Return $f_{server}$

\end{algorithmic}
\vspace{-1.3mm}
\end{algorithm}

\subsection{Collaborative Style Training Strategy}\label{sec32}
Unlike traditional learning strategies, we propose a novel collaborative training strategy for new and old styles, aiming to achieve rapid adaptation to new styles and continuous utilization of positive styles. As shown in Fig .\ref{fig2} (b), our strategy consists of two branches: (1) New style adaptation branch: This branch adjusts the style statistics of the original images to generate diverse data, and then uses the client-global model for rapid learning and adaptation. (2) Positive style continuous utilization branch: This branch focuses on continuously learning and utilizing verified positive styles. In each iteration, we leverage the memory mechanism’s ability to remember and reinforce the positive styles learned. This helps the model maintain a rich style knowledge system during long-term training, thereby achieving more robust and sustained generalization performance improvement.


\textbf{The new style adaptation branch.} To ensure that the client models can still achieve sufficient generalization ability with limited data, it is essential for them to continuously learn new styles. We refer to the stylization method proposed in \cite{52} and input the original image $x_{k}^{i}$ into the stylization network at the beginning of each iteration to generate a new stylized image $\hat{x}_{k}^{i}$ like Eq. \ref{eq2}. Then, we download the global model from the server to the client, referred to as the client-global model $f_{Client-G}$. Compared to the client-local model, the client-global model possesses richer global generalization knowledge. Coupled with the rapid convergence characteristic of the cross-entropy loss, it can achieve faster learning and adaptation to new styles. Meanwhile, we also employ the widely used triplet loss. Therefore, the training of the new style adaptation branch on the $k-th$ client can be expressed as:


\begin{equation}
    \begin{aligned}
        L_{NS}^{k} &= L_{CE}(cls_{k}(f_{Client-G}^{k}(\hat{x}_{k}^{i})),  y_k^i) \\
        &+ L_{Tri}(f^{k}_{Client-G}(\hat{x}_{k}^{i}, y_k^i)),
    \end{aligned}
    \label{eq5}
\end{equation}

among them, $\text{cls}_k$ is a classifier maintained by the client $k$ that possesses local knowledge, used for precise classification, $y_k^i$ is the label corresponding to $x_k^i$, $L_{CE}$ stands for calculation of cross entropy and $L_{Tri}$ stands for calculation of triplet loss.


\begin{table*}[th!]
\centering
\renewcommand{\arraystretch}{1.25}
\caption{Comparison of different methods under \textit{protocol-1}. M: Market1501, C2: CUHK02, C3: CUHK03, MS: MSMT17. Average represents the average performance over three unseen domains.}
\label{table1}
\begin{tabular}{c|c|c||cc|cc|cc|cc}
\Xhline{1.5pt}
\rowcolor{mygray}
 & & & \multicolumn{2}{c|}{\textbf{MS+C2+C3→M}} & 
\multicolumn{2}{c|}{\textbf{M+C2+C3→MS}} & \multicolumn{2}{c|}{\textbf{MS+C2+M→C3}} &
\multicolumn{2}{c}{\textbf{Average}}
\\
\cline{4-11}
\multirow{-2}{*}{\textbf{Category}}\cellcolor{mygray}& \multirow{-2}{*}{\textbf{Methods}}\cellcolor{mygray}& \multirow{-2}{*}{\textbf{Reference}}\cellcolor{mygray}&\cellcolor{mygray} mAP &\cellcolor{mygray} rank-1 &\cellcolor{mygray} mAP &\cellcolor{mygray} rank-1 &\cellcolor{mygray} mAP &\cellcolor{mygray} rank-1 &\cellcolor{mygray} mAP &\cellcolor{mygray} rank-1\\ 
\hline
\hline
\multirow{3}{*}{Federated Learning} 
& SCAFFOLD\cite{56} & ICML 2020 & 26.0 & 50.5 & 5.3 & 15.8 & 22.9 & 26.0 & 18.1 & 30.8 \\
& \cellcolor{mygray} MOON\cite{58} & \cellcolor{mygray} CVPR 2021 & \cellcolor{mygray}26.8  & \cellcolor{mygray}51.1 &\cellcolor{mygray} 4.8 & \cellcolor{mygray}14.5 & \cellcolor{mygray}20.9 &\cellcolor{mygray} 22.5 &\cellcolor{mygray}17.5 & \cellcolor{mygray}29.4\\
& FedProx\cite{57} & MLSys 2020 & 29.3 & 53.8 & 5.8 & 17.4 & 19.1 & 17.7 & 18.1 & 29.6 \\
\cline{1-3}\cline{4-11}
\multirow{2}{*}{Domain Generalization} 
& \cellcolor{mygray} MixStyle\cite{29} & \cellcolor{mygray} ICLR 2021 &\cellcolor{mygray} 31.2 & \cellcolor{mygray}53.5 &\cellcolor{mygray} 5.5 &\cellcolor{mygray} 16.0 &\cellcolor{mygray} 28.6 & \cellcolor{mygray}31.5 &\cellcolor{mygray}21.8 &\cellcolor{mygray} 33.7\\
& CrossStyle\cite{61} & ICCV 2021& 35.5 & 59.6 & 4.6 & 14.0 & 27.8 & 28.0 & 22.6 & 33.9\\
\cline{1-3}\cline{4-11}
\multirow{2}{*}{Federated-ReID} 
& \cellcolor{mygray} FedReID\cite{20} & \cellcolor{mygray} AAAI 2021 & \cellcolor{mygray}30.1 & \cellcolor{mygray}53.7 &\cellcolor{mygray} 4.5 & \cellcolor{mygray}13.7 & \cellcolor{mygray}26.4 & \cellcolor{mygray}26.5&\cellcolor{mygray}20.3&\cellcolor{mygray} 31.3 \\
& FedPav\cite{49} & MM 2020& 25.4 & 49.4 & 5.2 & 15.5 & 22.5 & 24.3 &17.7 & 29.7\\
\cline{1-3}\cline{4-11}

\multirow{1}{*}{DG-ReID} 
&\cellcolor{mygray} SNR\cite{41} &\cellcolor{mygray} CVPR 2020 &\cellcolor{mygray} 32.7 & \cellcolor{mygray}59.4 &\cellcolor{mygray} 5.1 &\cellcolor{mygray} 15.3 &\cellcolor{mygray} 28.5 & \cellcolor{mygray}30.0 &\cellcolor{mygray}22.1 &\cellcolor{mygray} 34.9\\
\cline{1-3}\cline{3-11}

\multirow{2}{*}{FedDG-ReID}
& DACS\cite{52} & AAAI2024& 36.3 & 61.2 & 10.4 & 27.5 & 30.7 & \textbf{34.1} & 25.8 & 40.9 \\
&\cellcolor{tab1huang} \textbf{SSCU (ours)}  & \cellcolor{tab1huang} This Paper & \cellcolor{tab1huang}{\textbf{39.5}} & \cellcolor{tab1huang}{\textbf{66.4}} & \cellcolor{tab1huang}{\textbf{11.9}} & \cellcolor{tab1huang}{\textbf{32.3}} &\cellcolor{tab1huang} {\textbf{32.8}}& \cellcolor{tab1huang}{\textbf{34.1}}& \cellcolor{tab1huang}{\textbf{28.1}}&\cellcolor{tab1huang} {\textbf{44.3}}\\
\hline
\Xhline{1.5pt}
\end{tabular}
\vspace{-2mm}
\end{table*}

\textbf{Positive style continuous utilization branch.} Existing learning strategies do not take into account the issue that not all styles contribute to the improvement of model generalization performance, and therefore cannot achieve continuous and effective utilization of positive styles while learning new ones. To address this issue, we design a continuous utilization branch for positive styles, coupled with the previously introduced dynamic style memory, to enable effective training of client models using positive styles continuously. Specifically, in this branch, both the client-local model and the client-global model are trained using the untransformed original images $x_k^i$ as input, and optimized using the style memory recognition loss introduced earlier (Eq. \ref{eq4}):


\begin{align}
L_{PS\_L}^k(f^k_{Client-L},x_k^i,M_k^i) &= L_{id}(f^k_{Client-L},x_k^i,M_k^i), \label{eq6}\\
L_{PS\_G}^k(f^k_{Client-G},x_k^i,M_k^i) &= L_{id}(f^k_{Client-G},x_k^i,M_k^i), \label{eq7}
\end{align}

\begin{equation}
    \begin{aligned}
       L_{PS}^k(f^k_{Client-L},f^k_{Client-G},x_k^i,M_k^i) &= L_{PS\_L}^k(f^k_{Client-L},x_k^i,M_k^i)\\
       &+ L_{PS\_G}^k(f^k_{Client-G},x_k^i,M_k^i),
    \end{aligned}
    \label{eq8}
\end{equation}

where $f_{Client-L}^k$ is the client-local model of client $k$. By applying recognition loss to the client-local model, it helps to maintain specific domain knowledge, thereby assisting the style transformation model in generating higher-quality data \cite{52}. Meanwhile, applying recognition loss to the client-global model enables continuous utilization and reinforcement of positive styles that have been quickly adapted, further enriching the model’s generalization knowledge.


\subsection{Subsequent Learning}

\textbf{Model upload and aggregation.} After completing local training, each client uploads the feature extractor part of its client-global model to the central server. A data-volume-weighted aggregation strategy is employed for fusion, ensuring that clients with more abundant data have a greater “say” in the global model aggregation. 


\textbf{Continuous training and final testing.} The clients and the server iteratively perform model download, local training, model aggregation, model redistribution until training convergence. Our entire methodological process is illustrated in Alg. \ref{alg1}.


\section{Experiments}

\subsection{Experimental Settings}

\textbf{Datasets.} We conduct experiments on four large-scale datasets: CUHK02, CUHK03, MSMT17, and Market1501. For simplicity, we refer to them as C2 (CUHK02), C3 (CUHK03), MS (MSMT17), and M (Market1501). In order to make a comprehensive comparison of the generalization performance on the source and target domains with existing works, we adopt three evaluation schemes:


\textit{protocol-1:} Under this protocol, we adopt a leave-one-out strategy, where each time one dataset is selected from the four datasets mentioned above as the test set, and all the remaining datasets are used as the training set.


\textit{protocol-2:} is a supplement to \textit{protocol-1}. For the two cases in \textit{protocol-1} where MS is used as the training dataset, we reduce the number of source domains by one to test the generalization performance of our method with fewer source domains(MS is always used for training).


\textit{protocol-3:} This protocol we choose C2, C3, and M as the training datasets, and separately use the test data from one of these datasets to test on the source domains.



\textbf{Implementation Details.} We treat each source domain as a client, using the same network as DACS \cite{52} as our backbone. The number of training epochs is set to 60, the batch size is set to 64, the initial learning rate is set to 1e-3, weight decay is set to 5e-4, momentum is set to 0.9, and MultiStepLR is used as the learning rate scheduler with milestones set at the 20th and 40th epochs. We use python (3.9) and PyTorch (2.1.0) to train on two Nvidia GeForce RTX-2080Ti GPUs.



\begin{table}[th!]
\caption{Comparison of different methods under \textit{protocol-2}. Under this protocol, we use fewer source domains to further compare generalization performance.} 
\label{table2} 
\resizebox{\linewidth}{!}{
\begin{tabular}{c|cc|cc|cc} 
\Xhline{1.5pt}
\rowcolor{orange}
\multicolumn{1}{c|}{} & \multicolumn{2}{c|}{MS+C3→M} & \multicolumn{2}{c|}{MS+C2→M} & \multicolumn{2}{c}{MS+C2+C3→M} \\
\cline{2-7}
\rowcolor{orange}
\multicolumn{1}{c|}{\cellcolor{orange}\multirow{-2}{*}{\textbf{Methods}}} & mAP & rank-1 & mAP & rank-1 & mAP & rank-1 \\ \hline
FedPav & 27.5 & 51.5 & 24.8 & 48.5 & 25.4 & 49.4 \\
\rowcolor{orangegray}
FedReID    &  31.0 & 55.0  & 28.1  & 52.4  &  30.1 & 53.7 \\
DACS    & 33.2  & 58.1  &  30.3 & 56.3  &  36.3 & 61.2 \\
\rowcolor{tab1huang}
\textbf{SSCU (ours)}   & \textbf{36.7} & \textbf{62.8} & \textbf{34.8} & \textbf{62.7}& \textbf{39.5} & \textbf{66.4}\\ \hline

\rowcolor{orange}
\multicolumn{1}{c|}{} & \multicolumn{2}{c|}{MS+M→C3} & \multicolumn{2}{c|}{MS+C2→C3} & \multicolumn{2}{c}{MS+C2+M→C3} \\
\cline{2-7}
\rowcolor{orange}
\multicolumn{1}{c|}{\cellcolor{orange}\multirow{-2}{*}{\textbf{Methods}}} & mAP & rank-1 & mAP & rank-1 & mAP & rank-1 \\ \hline
FedPav & 15.2 & 14.1 & 17.3 & 17.0 & 22.5 & 24.3 \\
\rowcolor{orangegray}
FedReID &  16.1 & 15.3&  21.8 & 20.4  &26.4   &  26.5 \\
DACS  & 18.2 & 17.7 & 22.9 & 23.5 & 30.7 & \textbf{34.1} \\ 
\rowcolor{tab1huang}
\textbf{SSCU (ours)}  & {\textbf{20.9}} & {\textbf{20.8}} & {\textbf{27.1}} & {\textbf{29.3}} & {\textbf{32.8}} & {\textbf{34.1}} \\ 
\Xhline{1.5pt}
\end{tabular}}
\vspace{-2mm}
\end{table}

\begin{table}[th!]
\caption{Comparison of different methods on \textit{protocol-3}. Under this protocol, we focus on the model's identification performance on the source domains, the most basic yet often overlooked capability by existing methods.}
\label{table3}
\resizebox{\linewidth}{!}{
\centering
\begin{tabular}{c|cc|cc|cc}
\Xhline{1.5pt}
\rowcolor{myblue}
\multicolumn{1}{c|}{} & \multicolumn{2}{c|}{M+C2+C3} & \multicolumn{2}{c|}{M+C2+C3} & \multicolumn{2}{c}{M+C2+C3} \\
\rowcolor{myblue}
\multicolumn{1}{c|}{}& \multicolumn{2}{c|}{$\rightarrow$ M} & \multicolumn{2}{c|}{$\rightarrow$ C2} & \multicolumn{2}{c}{$\rightarrow$ C3} \\
\cline{2-3} \cline{4-5} \cline{6-7}
\rowcolor{myblue}
\multicolumn{1}{c|}{\cellcolor{myblue}\multirow{-3}{*}{\textbf{Methods}}} & mAP & rank-1 & mAP & rank-1 & mAP & rank-1 \\
\hline
\textbf{FedProx} & 61.0 & 80.4 & 66.8 & 65.5 & 24.2 & 23.9 \\
\rowcolor{myllblue}
\textbf{FedPav} & 53.9 & 76.0 & 59.7 & 56.3 & 19.6 & 19.6 \\
\textbf{FedReID} & 71.8 & 87.6 & 82.9 & 82.8 & 44.0 & 44.9 \\
\rowcolor{myllblue}
\textbf{DACS} & 72.1 & 88.2 & 84.5 & 83.4 & 47.4 & 50.1 \\
\hline
\rowcolor{tab1huang}
\textbf{SSCU (ours)}  & {\textbf{73.0}} & {\textbf{88.7}} & {\textbf{84.9}} & {\textbf{83.9}} & {\textbf{50.4}} & {\textbf{53.2}} \\
\Xhline{1.5pt}
\end{tabular}}
\vspace{-2mm}
\label{tbl:methodscomparison}
\end{table}

\begin{table}[th!]
\centering
\renewcommand{\arraystretch}{1.25}
\caption{Ablation study on new style adaptation branch (NSA) and positive style continuous utilization branch (PSCU).}
\label{tab4}

\resizebox{\linewidth}{!}{
\begin{tabular}{cc|cc|cc}
\Xhline{1.5pt}
 \rowcolor{gree}
  \multicolumn{2}{c|}{} & \multicolumn{2}{c|}{\textbf{MS+C2}} & \multicolumn{2}{c}{\textbf{M+C2}} \\
 \rowcolor{gree}
 \multicolumn{2}{c|}{\multirow{-2}{*}{\textbf{Attributes}}} & \multicolumn{2}{c|}{\textbf{+C3→M}} & \multicolumn{2}{c}{\textbf{+C3→MS}}  \\

\hhline{>{\arrayrulecolor{black}}------}
\rowcolor{gree}
  NSA & PSCU & mAP & rank-1 & mAP & rank-1  \\
\hline
\hline
 $\times$ & $\times$ & 20.8 & 51.9 & 6.1 & 20.7 \\
 \rowcolor{greegray}
 $\times$ & \checkmark & \makecell{$33.6$ \\ \textcolor{red!40}{$\uparrow+12.8$}} & \makecell{$61.3$ \\ \textcolor{red!40}{$\uparrow+9.4$}} & \makecell{$7.5$ \\ \textcolor{red!40}{$\uparrow+1.4$}} & \makecell{$22.6$ \\ \textcolor{red!40}{$\uparrow+1.9$}} \\
 
  \checkmark  & $\times$ & \makecell{$36.2$ \\ \textcolor{red!40}{$\uparrow+15.4$}} & \makecell{$63.9$ \\ \textcolor{red!40}{$\uparrow+12.0$}} & \makecell{$7.9$ \\ \textcolor{red!40}{$\uparrow+1.8$}} & \makecell{$24.7$ \\ \textcolor{red!40}{$\uparrow+4.0$}} \\
\rowcolor{greegray}
 \checkmark & \checkmark & \makecell{\textbf{39.5}  \\ \textcolor{red!40}{$\uparrow+18.7$}}& \makecell{\textbf{66.4} \\ \textcolor{red!40}{$\uparrow+14.5$}} &\makecell{\textbf{11.9} \\ \textcolor{red!40}{$\uparrow+5.8$}} & \makecell{ \textbf{32.3} \\ \textcolor{red!40}{$\uparrow+11.6$}}  \\
\Xhline{1.5pt}
\end{tabular}}
\vspace{-6mm}
\end{table}

\subsection{Comparisons With State-of-the-art Methods}

We conduct comparisons with SOTA methods under three different protocols to demonstrate the effectiveness of our approach.


\textit{protocol-1:} As shown in Tab. \ref{table1}, the methods we compare can be categorized into five classes: (1) First are the classical federated learning algorithms, such as SCAFFOLD \cite{56}, FedProx \cite{57} and MOON \cite{58}. (2) Next are domain generalization methods based on styling approaches, with the most classical being MixStyle \cite{29} and CrossStyle \cite{61}, which we deploy directly on the clients to generate diverse data for local training. (3) Following this are the representative federated ReID methods, such as FedPav \cite{49} and FedReID \cite{20}. (4) Subsequently, there are single-source domain generalization ReID methods without privacy constraints, such as SNR \cite{41}, where we deploy each module after every layer of ResNet. (5) Finally, there are federated domain generalization ReID methods, such as DACS \cite{52}. The experimental data shows that our method achieves state-of-the-art performance. Specifically, for “MS+C2+C3→M”, the mAP reaches 39.2\%, and Rank-1 reaches 66.4\%, which is at least \textcolor{red!40}{$\uparrow2.9\%$} and \textcolor{red!40}{$\uparrow5.2\%$} ahead of other methods respectively. For “M+C2+C3→MS”, the mAP reaches 11.9\%, and Rank-1 reaches 32.3\%, which is at least \textcolor{red!40}{$\uparrow1.5\%$} and \textcolor{red!40}{$\uparrow4.8\%$} ahead of other methods respectively. For “M+C2+MS→C3”, the mAP reaches 32.8\%, which is at least \textcolor{red!40}{$\uparrow2.1\%$} higher than other methods, and Rank-1 reaches 34.1\%, on par with the best-performing method.

\textit{protocol-2:} As shown in Tab. \ref{table2}, for settings "MS+C2+C3→M" and "M+C2+MS→C3", we respectively remove C2, C3, and M from the source domains to compare our method with the state-of-the-art methods under the condition of having fewer source domain data. We compare our method with those that still perform excellently in this field, such as the FedPav \cite{49}, FedReID \cite{20}, and DACS \cite{52}. It can be observed that even in the face of the challenge of reduced source domain data, our method still outperforms the SOTAs.

\textit{protocol-3:} As shown in Tab. \ref{table3}, we deployed the models trained under the source domain setting of \textit{protocol-1} back to each source domain to test the recognition capability of our method on the source domains and also compared it with the state-of-the-art methods. The experimental results show that our method not only achieves excellent generalization performance but also ensures the optimal recognition performance on the source domains.

\begin{figure*}[t]
    \centering
    \includegraphics[width=\textwidth]{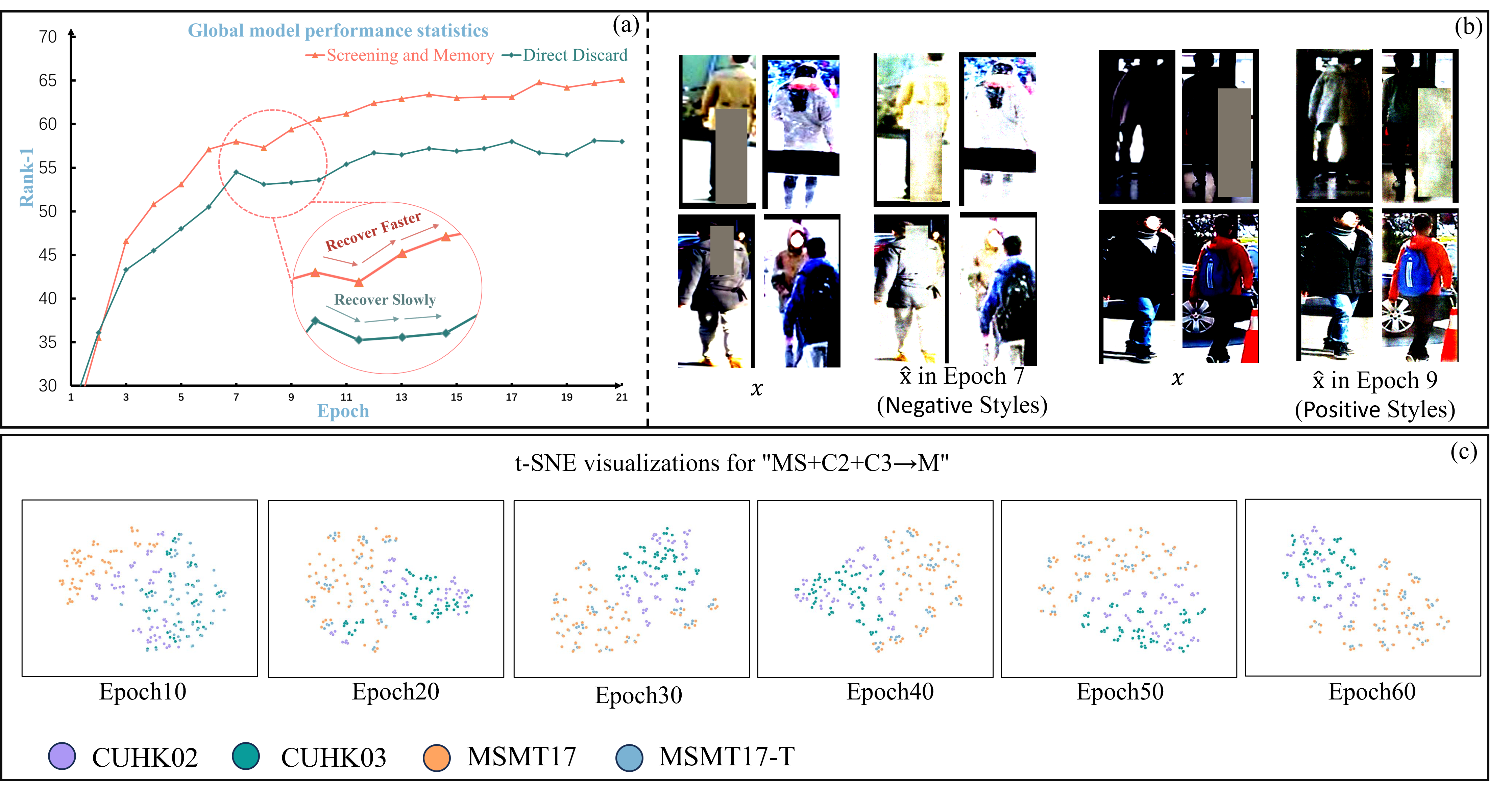}
    \caption{(a) Performance statistics of the method that directly discards generated styles and the method that screens and memorizes styles. (b) Display of positive and negative styles during the training process. (c) $t-SNE$ visualization of “MS+C2+C3→M”. MSMT17-T: Style data generated from the MSMT17 dataset.}    
    \label{fig3} 
    \vspace{-2mm}
\end{figure*}


\subsection{Ablation Study}

\textbf{Effectiveness of the NSA and PSCU.} As previously mentioned, our CST strategy consists of two branches: the New Style Adaptation (NSA) branch, and the Positive Style Continuous Utilization (PSCU) branch. We explored the effects of these two branches separately to demonstrate the effectiveness of our strategy. "Baseline" indicates that we directly trained the client-global model using the original images and cross-entropy loss during the local training phase. "Baseline+NSA" means that we trained both the client-local and client-global models using the original images, while also training the client-global model with style-transformed images, but without introducing the memory mechanism. Therefore, the classification loss used was the cross-entropy loss function. "Baseline+PSCU" indicates that for each round of generated style images, they were directly memorized in the memory without being used for the training of the client-global model in the current round. The classification loss was our proposed style memory identification loss. "Baseline+NSA+PSCU" is the method we introduced earlier. All settings were the same as our basic training phase. As shown in the Tab. ~\ref{tab4}, our training pipeline shows significant improvement. Both branches contribute to the improvement of the model's generalization performance, and the best results are achieved when they are used in conjunction.

\subsection{Visualization}
During the implementation of the "MS+C2+C3→M" experiment, we statistically analyze the performance changes of our method when applying the screening and memorizing strategy (introduced in the previous text) for styles and the direct discarding method (used in existing approaches), and plotted the results as line graphs for a better understanding of our solution. As shown in Fig. \ref{fig3}(a), both methods experienced a severe performance drop after the completion of the seventh round of training. However, due to our screening and memorizing strategy, the model could continuously benefit from positive styles, thus recovering more quickly from the impact of the performance decline. In Fig. \ref{fig3}(b), we present the negative styles generated in the seventh round, where it can be observed that the transformed images have lost a significant amount of detail, which is detrimental to model optimization. Correspondingly, we showcase the positive styles generated in the ninth round, where it is clearly evident that these images retain detailed information while undergoing style transformation, and following the ninth round of training for both methods, there was an enhancement in their performance. In the $t-SNE$ visualization shown in Fig. \ref{fig3}(c), it can be clearly observed that in the initial stage of training, the feature points of different datasets are distributed relatively dispersedly in the feature space, without forming distinct distinctions. As the training progresses step by step, the feature points gradually gather towards the centers of their respective datasets, forming tighter and more defined clusters, and gradually stabilize. It is worth noting that during the analysis of the CHUK02 and CUHK03, it was found that these two datasets have a certain similarity in content, resulting in some features being distributed relatively close.

\section{Conclusion}
In this paper, we propose a Style Screening and Continuous Utilization (SSCU) framework for federated generalized person re-identification. This method not only achieves precise screening of positive styles but also ensures their maximum utilization through an efficient memory mechanism. Specifically, our approach first introduces a generalization gain-guided dynamic style memory for screening and memorization. To ensure the continuous utilization of these positive styles, we further design a style memory recognition loss function. Correspondingly, we devise a style collaborative training strategy to simultaneously learn newly generated styles and the positive styles stored in memory. The new style rapid adaptation branch facilitates quick adaptation to new style changes through the use of a client-global model. The positive style continuous utilization branch is responsible for fully leveraging the positive styles. Extensive experiments show the efficacy of our method.


\begin{acks}
This work was supported by the National Nature Science Foundation of China (No. 62376201). This research was financially supported by funds from Key Laboratory of Social Computing and Cognitive Intelligence (Dalian University of Technology), Ministry of Education (No.  SCCI2024YB02).
\end{acks}

\bibliographystyle{ACM-Reference-Format}
\bibliography{sample-base}

\end{document}